\documentclass[sigconf,screen]{acmart}

\AtBeginDocument{%
  \providecommand\BibTeX{{%
    \normalfont B\kern-0.5em{\scshape i\kern-0.25em b}\kern-0.8em\TeX}}}

\copyrightyear{2022}
\acmYear{2022}
\setcopyright{acmcopyright}\acmConference[WWW '22]{Proceedings of the ACM Web Conference 2022}{April 25--29, 2022}{Virtual Event, Lyon, France}
\acmBooktitle{Proceedings of the ACM Web Conference 2022 (WWW '22), April 25--29, 2022, Virtual Event, Lyon, France}
\acmPrice{15.00}
\acmDOI{10.1145/3485447.3512122}
\acmISBN{978-1-4503-9096-5/22/04}

\usepackage{xspace}
\usepackage{multirow}
\usepackage{multicol}
\usepackage{booktabs}
\usepackage{xcolor}
\usepackage{colortbl}
\begin{document}

\title{Evidence-aware Fake News Detection with Graph Neural Networks}
\author[ ]{Weizhi Xu$^{1,2,*}$, Junfei Wu$^{3,*}$, Qiang Liu$^{1,2}$, Shu Wu$^{1,2,\dagger}$, Liang Wang$^{1,2}$}
\makeatletter
\def\authornotetext#1{
	\g@addto@macro\@authornotes{%
	\stepcounter{footnote}\footnotetext{#1}}%
}
\makeatother

\authornotetext{The first two authors contributed equally to this work.}
\authornotetext{Corresponding author.}

\affiliation{%
	\institution{$^1$CRIPAC, NLPR, Institute of Automation, Chinese Academy of Sciences}
	\institution{$^2$School of Artificial Intelligence, University of Chinese Academy of Sciences \quad $^3$Beijing Institute of Technology}
    \country{}
}

\email{weizhi.xu@cripac.ia.ac.cn,            junfei.wu@bit.edu.cn, {qiang.liu,shu.wu,wangliang}@nlpr.ia.ac.cn}

\def\authors{Weizhi Xu, Junfei Wu, Qiang Liu, Shu Wu, Liang Wang}

\newcommand{\themodel}{GET\xspace}


\begin{abstract}
  The prevalence and perniciousness of fake news has been a critical issue on the Internet, which stimulates the development of automatic fake news detection in turn. In this paper, we focus on the evidence-based fake news detection, where several evidences are utilized to probe the veracity of news (i.e., a claim). Most previous methods first employ sequential models to embed the semantic information and then capture the claim-evidence interaction based on different attention mechanisms. Despite their effectiveness, they still suffer from two main weaknesses. Firstly, due to the inherent drawbacks of sequential models, they fail to integrate the relevant information that is scattered far apart in evidences for veracity checking. Secondly, they neglect much redundant information contained in evidences that may be useless or even harmful. To solve these problems, we propose a unified \textbf{\underline{G}}raph-based s\textbf{\underline{E}}mantic s\textbf{\underline{T}}ructure mining framework, namely \themodel in short. Specifically, different from the existing work that treats claims and evidences as sequences, we model them as graph-structured data and capture the long-distance semantic dependency among dispersed relevant snippets via neighborhood propagation. After obtaining contextual semantic information, our model reduces information redundancy by performing graph structure learning. Finally, the fine-grained semantic representations are fed into the downstream claim-evidence interaction module for predictions. Comprehensive experiments have demonstrated the superiority of \themodel over the state-of-the-arts.
\end{abstract}





\maketitle

\section{introduction}
\begin{figure}[t]
   \begin{center}
   \includegraphics[width=0.47\textwidth]{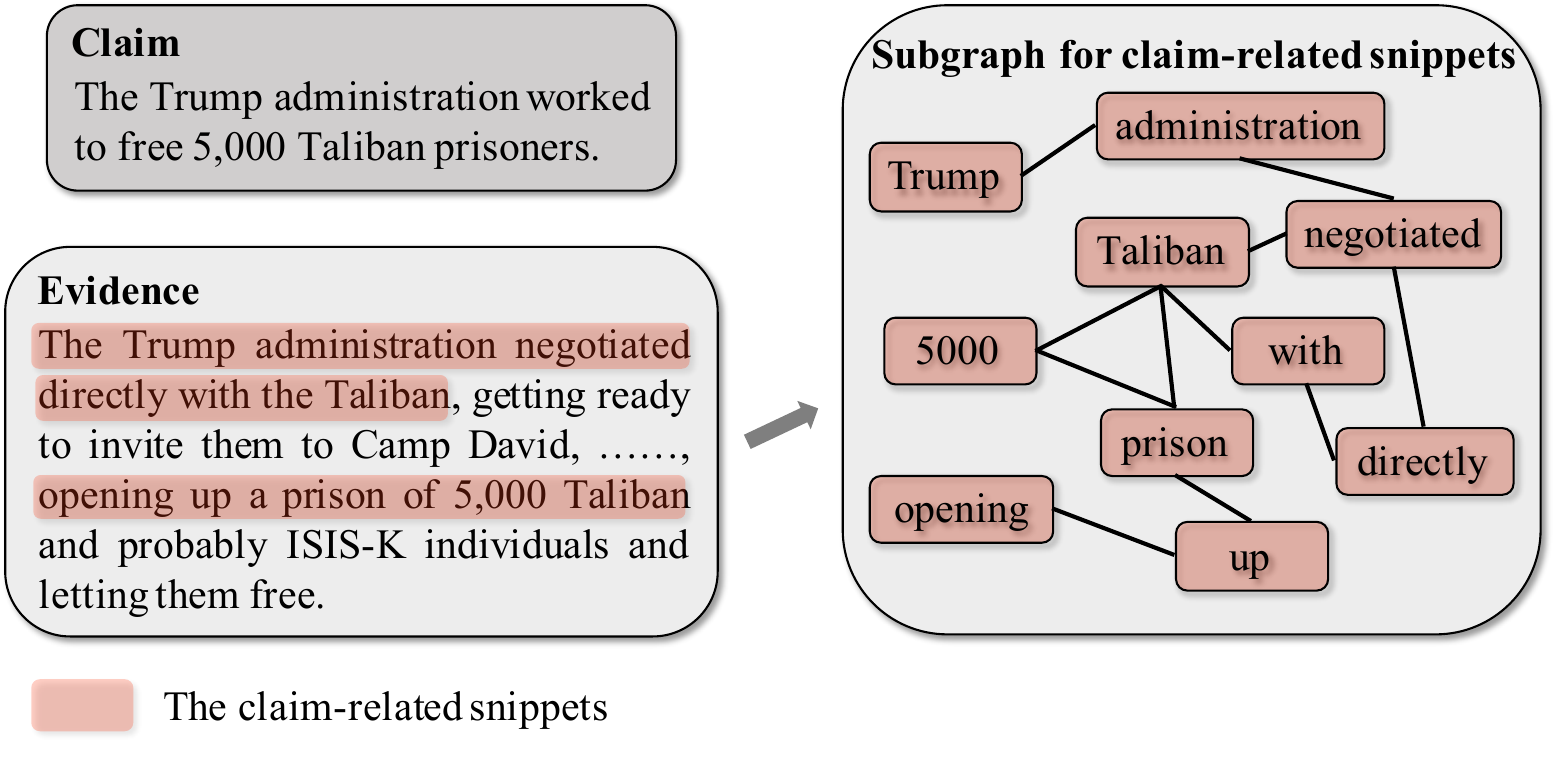}
   \end{center}
   \caption{A toy example where a claim and its relevant evidence are given. Two significant snippets for verifying the claim are highlighted (``.....'' represents that we omit several sentences for conciseness). The right graph is constructed according to the highlighted snippets. Such two snippets have a long distance in the plain text while they are pulled close on the constructed semantic graph via the shared keyword ``Taliban''. Besides, there is much redundant information (texts except the highlighted parts), which is useless for claim verification.}
   \label{fig:example}
\end{figure}
Fake news, which is always fabricated by making some minor changes to the correct statement, is highly deceptive and indistinguishable. The widespread of fake news in diverse domains, such as politics \cite{Allcott2017SocialMA} and public health \cite{Naeem2020TheC}, has posed a huge threat to web security and human society. Therefore, the research on automatic fake news detection is challenging but in demand.

Generally, previous methods could be roughly categorized into two groups, i.e., pattern-based approaches and evidence-based approaches \cite{Sheng2021IntegratingPA}. The former methods regard the fake news detection as a feature recognition task, where language models are employed to verify the veracity of news solely according to the text pattern, e.g., writing styles. 
However, pattern-based methods usually suffer from the poor generalization and interpretability. The latter approaches model the task as a reasoning process, where external evidences are provided to probe the veracity of a claim. Models are required to discover and integrate useful information in given evidences for claim verification.

In this paper, we concentrate on the evidence-based pipeline. Existing methods usually follow a two-step paradigm: 1) they first capture the semantics of claims and evidences separately. 2) Next, they model the claim-evidence interaction to explore the semantic coherence or conflict for more accurate and interpretable verdict. To name a few representative models, the pioneering work DeClarE \cite{popat2018declare} utilizes bidirectional LSTMs to model textual features, followed by a word-level attention mechanism to capture the claim-evidence interaction. HAN \cite{ma2019sentence} further considers the sentence-level interaction to explore more general semantic coherence. To obtain multi-level semantic interaction, some recent works \cite{vo2021hierarchical, wu2021unified} employ hierarchical attention networks. 

Nevertheless, existing work focuses on the specific design of different interaction models (the second step) while neglecting exploring fine-grained semantics of claims and evidences (the first step). To be specific, we argue that there are two main weaknesses in previous methods. 

Firstly, the complex, long-distance semantic dependency is less explored. Taking Figure \ref{fig:example} as an example, two highlighted snippets are separated by plenty of words, which induces a long distance between them. Such snippets both contain important information for verifying the claim, i.e., the subject ``The Trump administration'' and the action ``opening up a prison of 5,000 Taliban''. Therefore, fusing the information is indispensable and beneficial for claim veracity prediction. However, the long-distance semantic dependency between such information is hard to be captured due to the inherent drawbacks of sequential models utilized in previous methods. 

Secondly, existing methods neglect the redundant information involved in semantics. Such redundancy is useless or even harmful for fake news detection, e.g., as depicted in Figure \ref{fig:example}, a large number of text segments, such as ``getting ready to invite them to Camp David'', have no substantial contribution to the news veracity checking. Though previous models employ attention mechanisms to reduce the effect of unrelated words, these irrelevant texts are still preserved, which may introduce noises to the downstream claim-evidence interaction, deteriorating the final performance of veracity checking. 
An intuitive solution is to discard words with low attentive scores based on previous methods. However, they compute the score for each word independently, ignoring the complex semantic structure among words.  We argue that it is significant to modeling the redundancy with rich semantic structural information, as the redundancy is not only related to the self-information, but also induced by its contexts, e.g., if a claim can be verified by a snippet in an evidence, the snippet’s context will be redundant.

To tackle the aforementioned problems, we propose a unified \textbf{\underline{G}}raph-based s\textbf{\underline{E}}mantic s\textbf{\underline{T}}ructure mining framework, namely \themodel for exploring fine-grained semantics. Specifically, modeling sequential data as graphs has benefited many tasks, such as text classification \cite{Yao2019GraphCN, texting} and sequential recommendation \cite{srgnn}, owing to its capability of capturing long-distance structural dependency. To this end, we utilize graph structure to model both claims and evidences, where nodes indicate words and edges represent the co-occurence between two words. Thereafter, the dispersed claim-related snippets are pulled close on graphs, thus the useful information could be better fused via neighborhood propagation. For example, in Figure \ref{fig:example}, after constructing the graph for two highlighted snippets distant from each other in plain texts, they are pulled close via the shared keyword ``Taliban'' so that the long-distance semantic dependency can be captured. Moreover, to alleviate the negative impact of redundant information, within our graph-based framework, we treat the redundancy mitigation as a graph structure learning process, where unimportant nodes are discarded according to complex semantic structures, i.e., both self-features (node attributes) and their contexts (graph topology). 
To date, our graph-based framework has captured the fine-grained semantics via long-distance dependency modeling and redundancy mitigation. Based on such semantics, we can apply the widely used attention mechanism in previous work to readout node features and form the claim- and evidence-level representations, followed by claim-evidence interactions to integrate information for the final veracity prediction.

Our main contributions can be summarized as follows:
\begin{itemize}
    \item We model claims and evidences as graph-structured data and design a graph-based framework to explore the complex semantic structure. To the best of our knowledge, this is the first work to propose a unified graph-based method for evidence-based fake news detection.
    \item We introduce a simple and effective graph structure learning approach for redundancy mitigation. By capturing long-distance semantic dependency and reducing redundancy, we obtain the fine-grained semantics, which can boost the performance of downstream interaction models. 
    \item Comprehensive experiments are conducted to verify the effectiveness of \themodel, where the results demonstrate its superiority.
\end{itemize}
\section{related work}

\subsection{Graph Neural Networks}
Graph neural networks (GNNs) learn the node representation by gathering information from the neighborhood, i.e., neighborhood propagation/aggregation. Current GNNs can be roughly divided into two groups, namely spectral approaches \cite{Defferrard2016ConvolutionalNN, Kipf2017SemiSupervisedCW} and spatial approaches \cite{Velickovic2018GraphAN, Hamilton2017InductiveRL}. Owing to the capability of capturing long-distance structural relationship on graphs, GNNs have been widely utilized and achieved satisfactory performance in several tasks, such as recommender system \cite{srgnn, Chen2020HandlingIL, Zhang2021MiningLS}, text classification \cite{Yao2019GraphCN, texting}, and sentiment analysis \cite{Wang2020RelationalGA, Li2021DualGC}. 
Recently, researchers have observed that graphs inevitably contain noises that may deteriorate the training of GNNs \cite{Jin2020GraphSL}. To handle this problem, graph structure learning (GSL) is proposed, aiming to jointly learn an optimized graph structure and node embeddings. Existing GSL methods mainly fall into three groups \cite{zhu2021gsl}: 1) \emph{the metric-learning-based methods} where the adjacency matrices are built as metrics coupled with node embeddings. Therefore, the graph topology is updated with node embeddings being optimized. The metrics are mainly defined as the attention-based function \cite{jiang2019gsl, chen2020gsl, cosmo2020gsl} or kernel function \cite{li2018gsl, wu2018gsl}. 2) \emph{the probabilistic methods} assume that the adjacency matrix is generated by sampling from a specific probabilistic distribution \cite{franceschi2018gsl, franceschi2019gsl, Zhang2019gsl}. 3) \emph{the direct-optimized methods} treat the graph topology as learnable parameters that are updated together with task-specific parameters simultaneously, without depending on preset priors (node embeddings and distributions in the first two groups, respectively). The topology is optimized with the guidance of task-specific objectives (and some normalization constraints) \cite{Yang2019gsl, Jin2020GraphSL}. It is worth noting that existing graph pooling methods \cite{ying2018hierarchical, gao2019graph, lee2019self} could also be regarded as GSL algorithms, since the pooling target is to keep the most valuable nodes that preserve the graph structural information well, where the graph structure is optimized via merging or dropping nodes. Besides, GNNs are widely employed in the domain of fact verification, which have achieved promising performance \cite{Zhou2019GEARGE, Liu2020FinegrainedFV, Zhong2020ReasoningOS}. Though fact verification is similar to fake news detection on the task setting, the latter requires more fine-grained semantics since the texts consist of more redundancy.
\subsection{Fake News Detection}
Several fake news detection methods have been proposed in recent years, which can be roughly grouped into two categories.

The first is the pattern-based pipeline where models solely consider the text pattern involved in the news itself. Different work always focus on different kinds of patterns. \citet{popat2016} classify a claim as true or fake in accordance with stylistic features and the article stance. Besides, some researchers attempt to verify the truthiness via the feedback in social media, such as reposts, likes, and comments \cite{Yu2017ACA, volkova-etal-2017-separating, Vo2018, Benamira2019SemiSupervisedLA, Chandra2020GraphbasedMO, Jin2021TowardsFR, liu2018mining}. Recently, more attention has been paid to the emotional pattern mining, where they hold an assumption that there are probably obvious sentiment biases in fake news \cite{Ajao2019, Gia2019, zhang2021www}.

The second is the evidence-based pipeline where researchers propose to explore the semantic similarity (conflict) in claim-evidence pairs to check the news veracity. Evidences are usually retrieved from the knowledge graph \cite{vlachos-riedel-2015-identification} or fact-checking websites \cite{vlachos-riedel-2014-fact} by giving unverified claims as queries. DeClarE \cite{popat2018declare} is the first work to utilize evidences in fake news detection. They employ BiLSTMs to embed the semantics of evidences and obtain the claim's sentence-level representation via average pooling. Next, they introduce an attention-based interaction to compute the claim-aware score for each word in evidences. Similar to the pioneering work, the following methods utilize the sequential models to obtain the semantic embeddings, followed by attention mechanisms performed on different granularities. HAN \cite{ma2019sentence} compute the sentence-level coherence and entailment scores between claims and evidences. EHIAN \cite{wu2020evidence} employs the self-attention mechanism to obtain word-level interaction scores. Recent work \cite{vo2021hierarchical, wu2021unified, Wu_Rao_Sun_He_2021} hierarchically integrates both word-level and sentence-level interactions into the final representation for verification. In summary, they all employ sequential models to embed the semantics and apply attention mechanisms to capture the claim-evidence semantic relationship.

Different from the existing work, we propose a unified graph-based model, where the long-distance semantic dependency is captured via constructed graph structures and the redundancy is reduced by performing graph structure learning.

\section{method}

\begin{figure*}[t]
   \begin{center}
   \includegraphics[width=1\textwidth]{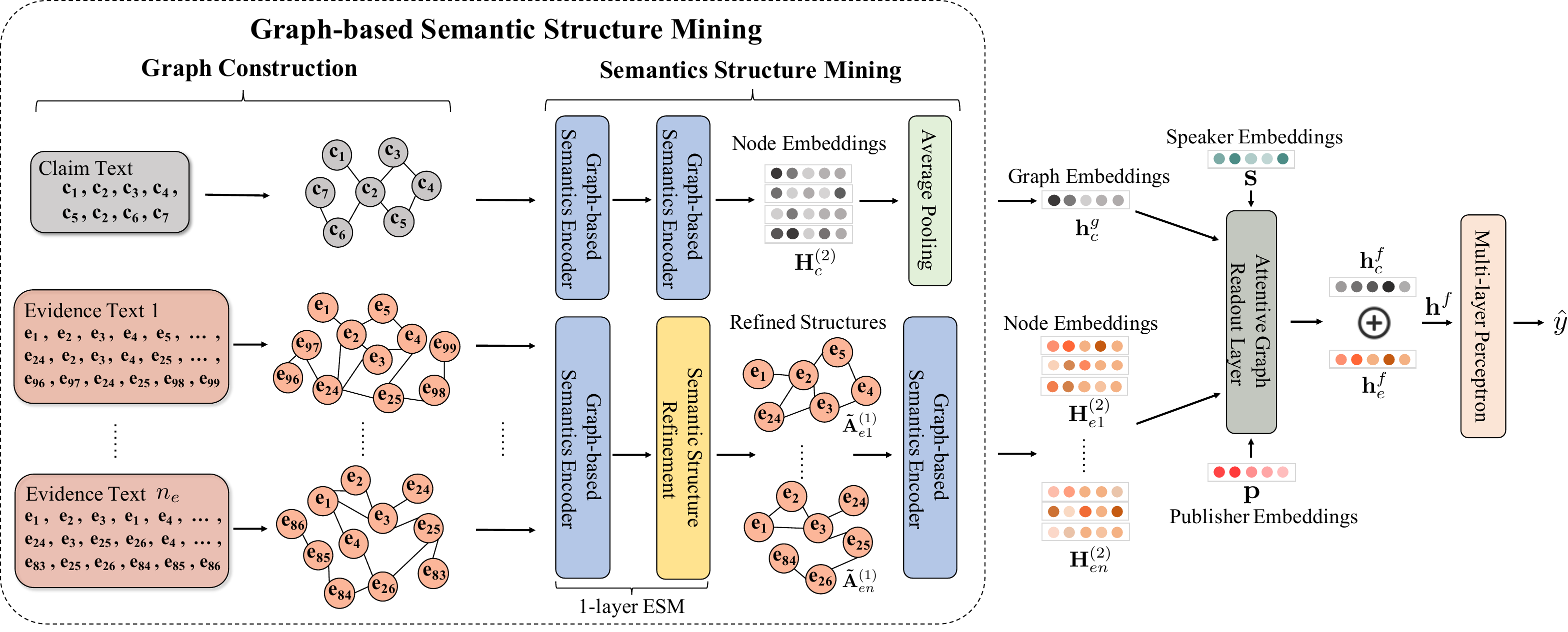}
   \end{center}
   \caption{The architecture of \themodel. The plain texts are first transformed into graphs using a sliding window (the window size is 2 in the figure). The same words repeatedly appear in texts are merged into one node. Next, we introduce graph-based semantics encoder to capture long-distance structural dependencies and generate high-order representations via neighborhood aggregation. Furthermore, the semantic structure refinement layer is proposed to generate optimized structures \(\{\mathbf{\tilde{A}}^{(1)}_{e1}, \ldots, \mathbf{\tilde{A}}^{(1)}_{en}\}\) for \(n\) evidences, where redundant nodes are discarded (The 1-layer ESM consists of a graph-based semantics encoder and a semantic structure refinement layer). Thereafter, the fine-grained semantics is obtained by performing neighborhood propagation on refined graphs. Finally, claim and evidence embeddings along with their speaker and publisher information are fed into the attentive graph readout layer to output the final prediction \(\hat{y}\).}
   \label{fig:model}
\end{figure*}

\subsection{Task Formulation}
Evidence-based fake news detection is a classification task, where the model is required to output the prediction of news veracity. Specifically, the inputs are a claim \(c\), several related evidences \(\mathcal{E} = \left\{e_1, e_2, \ldots, e_{n}\right\}\), and their corresponding speakers \(\mathbf{s} \in \mathbb{R}^{1 \times b}\) or publishers \(\mathbf{p} \in \mathbb{R}^{n \times b}\), where \(n\) is the number of evidences and \(b\) is the dimension of speaker and publisher embeddings. The output is the predicted probability of veracity \(\hat{y} = f\left(c, \mathcal{E}, \mathbf{s}, \mathbf{p}, \Theta \right)\), where \(f\) is the verification model and \(\Theta\) is its trainable parameters.

\subsection{The Proposed Model: \themodel}
In this part, we elaborate our unified graph-based model \themodel, which can be mainly separated into four modules: 1) \emph{Graph Construction}, 2) \emph{Graph-based Semantics Encoder}, 3) \emph{Semantic Structure Refinement}, and 4) \emph{Attentive Graph Readout Layer}. 
\subsubsection{Graph Construction}\quad
In order to capture the long-distance dependency of relevant information, we first convert the original claims and evidences to graphs. Like previous graph-based methods in other NLP tasks \cite{Yao2019GraphCN, grmm, ghrm, texting}, we use a fix-sized sliding window to screen out the connectivity for each word on graphs. In detail, the center words in every window will be connected with the rest of words in it (if connected, the corresponding entry in adjacency matrix is 1, otherwise 0), which captures the local context in center word's neighborhood. Furthermore, to model the long-distance dependency, we merge all the same words into one node on graph, which explicitly gathers their local contexts (e.g., the word \(e_2\) in evidence text 1 in Figure \ref{fig:model}). Therefore, several relevant snippets that scatter far apart is close on graphs, which can be explored via the high-order message propagation. In addition, the initial node representations are the corresponding word embeddings. Note that we also try to construct a graph in a fully-connected or semantic-similarity-based manner, but these two ways are all inferior to the sliding-window-based method, which may due to the redundant noises induced by the dense connection.

To ensure the numerical stability, we perform Laplacian normalization on adjacency matrices, denoted as \(\tilde{\mathbf{A}} = \mathbf{D}^{-\frac{1}{2}} (\mathbf{A}+\mathbf{I}) \mathbf{D}^{-\frac{1}{2}}\), where \(\mathbf{D}\) is the diagonal degree matrix (i.e., \(\mathbf{D}_{ii} = \sum_j \mathbf{A}_{ij}\)) and \(\mathbf{I}\) is the identical matrix. Finally, we denote the initial normalized adjacency matrices and node feature matrices of claim and evidence as \(\tilde{\mathbf{A}}^{(0)}_c \in \mathbb{R}^{N_c \times N_c}\), \(\tilde{\mathbf{A}}^{(0)}_e \in \mathbb{R}^{N_e \times N_e}\) and \(\mathbf{H}^{(0)}_c \in \mathbb{R}^{N_c \times d}\), \(\mathbf{H}^{(0)}_e \in \mathbb{R}^{N_e \times d}\), respectively. \(N_c\) and \(N_e\) is the number of nodes in initial claim and evidence graphs, \(d\) is the dimension of word embeddings.

Taking the established graph structures and node embeddings as inputs, we design a graph-based model to better capture complex semantics and obtain refined semantic structures.

\subsubsection{Graph-based Semantics Encoder}\quad
To mine the long-distance semantic dependency, we propose to utilize GNNs as the semantics encoder.
In particular, as we expect to adaptively keep a balance between self-features and the information of neighboring nodes, we employ graph gated neural networks (GGNN) to perform neighborhood propagation on both claim and evidence graphs, enabling nodes to capture their contextual information, which is significant for learning high-level semantics. Formally, it can be written as follows:
\begin{align}
    \label{eq:ggnn-s}
    \mathbf{a}_{i}&=\sum_{(w_{i}, w_{j}) \in \mathcal{C}} \mathbf{\tilde{A}}_{ij} \mathbf{W}_{a} \mathbf{H}_{j} \\
    \mathbf{z}_{i}&=\sigma\left(\mathbf{W}_{z} \mathbf{a}_{i}+\mathbf{U}_{z} \mathbf{H}_{i}+\mathbf{b}_{z}\right) \\
    \mathbf{r}_{i}&=\sigma\left(\mathbf{W}_{r} \mathbf{a}_{i}+\mathbf{U}_{r} \mathbf{H}_{i}+\mathbf{b}_{r}\right) \\
    \tilde{\mathbf{H}}_{i}&=\tanh \left(\mathbf{W}_{h} \mathbf{a}_{i}+\mathbf{U}_{h}\left(\mathbf{r}_{i} \odot \mathbf{H}_{i}\right)+\mathbf{b}_{h}\right) \\
    \mathbf{\hat{H}}_{i}&=\tilde{\mathbf{H}}_{i} \odot \mathbf{z}_{i}+\mathbf{H}_{i} \odot\left(1-\mathbf{z}_{i}\right) 
    \label{eq:ggnn-e}
\end{align}
where \(\mathcal{C}\) denotes the edge set, \(\mathbf{W}_*\), \(\mathbf{U}_*\), and \(\mathbf{b}_*\) are trainable parameters, which control the proportion of the neighborhood information and self-information. \(\sigma\) is the non-linear activation unit and we utilize the Sigmoid function in our model. For brevity, we denote Eq. (\ref{eq:ggnn-s}) - (\ref{eq:ggnn-e}) as \(\textbf{GGNN}(\mathbf{\tilde{A}}, \mathbf{H})\)\footnote{When generally describing the module that will be repeatedly utilized in the model, we omit the superscripts indicating layer number for brevity.}.

\subsubsection{Semantic Structure Refinement.}\quad
As evidences always contain redundant information that may mislead model to focus on unimportant features, it is beneficial to discover and filter out the redundancy, thus obtaining refined semantic structures. To this end, in our graph-based framework, we treat the redundancy mitigation as a graph structure learning process, whose aim is to learn the optimized graph topology along with better node representations. Previous GSL methods generally optimize the topology in three ways, i.e., dropping nodes, dropping edges, and adjusting edge weights. Since the redundancy information is mainly involved in words denoted as nodes in evidence graphs, we attempt to refine evidence graph structures via discarding redundant nodes, inspired by previous GSL methods \cite{lee2019self, chen2020gsl, Zhang2019gsl}. 

In particular, we propose to compute a redundancy score for each node, based on which we obtain a ranking list and nodes with the top-\(k\) redundancy scores will be discarded. The redundancy is not only related to the self-information contained in each node, but also induced by the contextual information, which is involved in the neighborhood on graphs. For example, if a claim can be verified by a snippet in an evidence, the rest of segments (including the snippet's context) will be redundant. Therefore, we utilize a 1-layer GGNN to compute the redundancy scores, which takes into account both self- and context-information in score computation. Mathematically, it can be formulated as:
\begin{align}
    \label{eq:ggnn-gsl}
    & \mathbf{s}_r = \textbf{GGNN}(\mathbf{\tilde{A}}, \mathbf{\hat{H}_e}\mathbf{W}_s) \\
    & idx = topk\_index(\mathbf{s}_r) \\
    & \mathbf{\tilde{A}}_{{idx}, :} = \mathbf{\tilde{A}}_{:, {idx}} = 0
    \label{eq:mask}
\end{align}
where \(\mathbf{W}_s \in \mathbb{R}^{d \times 1}\) is the trainable weights that project node representations into the 1-dimension score space. \(idx\) denotes the indices of node with top-\(k\) redundancy scores which are discarded by masking their degrees as 0 (c.f., Eq. (\ref{eq:mask})). Note that \(\textbf{GGNN}(\cdot)\) in Eq. (\ref{eq:ggnn-gsl}) does not share parameters with the semantics encoder due to their different targets. Besides, we only perform semantic structure refinement on evidences since claims are usually short (less than 10 words) so that the semantic structures are simple and unnecessary to be refined. 

Finally, we stack the semantic structure refinement layer over one semantics encoder to form a unified module, namely \emph{evidence semantics miner} (ESM in short), where the long-distance semantic dependency is captured and the redundant information is reduced. In general, we can stack \(T_R\) layers of ESM to refine the semantic structures \(T_R\) times, eventually followed by a semantics encoder to perform neighborhood propagation on refined semantic graphs, obtaining the fine-grained representations.

\subsubsection{Attentive Graph Readout Layer.}\quad
\label{sec:interaction module}
So far, we have obtained refined structures \(\mathbf{\tilde{A}}^{(T_{R})}_{e}\) for each evidence and fine-grained node embeddings \(\mathbf{H}^{(T_{E})}_{c}\), \(\mathbf{H}^{(T_{R}+1)}_e\) for claims and evidences separately\footnote{We omit the index subscript of evidences for brevity, as they are all fed into the same networks.}, where \(T_{R}\) and \(T_{E}\) are the numbers of ESM layer and semantics encoder layer of claim, respectively (\(T_{R} = 1\) and \(T_{E} = 2\) in Figure \ref{fig:model}). Next, to perform the claim-evidence interaction, we first need to integrate all node embeddings (word embeddings) into general graph embeddings (claim and evidence embeddings). Following previous work \cite{vo2021hierarchical}, we propose to obtain claim-aware evidence representations via the attention mechanism. In detail, we compute the attention score of the \(j\)-th word \(\mathbf{H}_{ej}\) in the refined evidence graph with the claim representation \(\mathbf{h}^g_c\). Thereafter, the evidence representation \(\mathbf{h}^{g}_{e}\) is obtained via weighted summation:

\begin{align}
    \mathbf{h}^g_c &= \frac{1}{l_c} \sum_{i=1}^{l_c} \mathbf{H}_{ci} \\
    \label{eq:attn-s}
    \mathbf{p}_j &= \tanh \left(\left[\mathbf{H}_{ej} ; \mathbf{h}^g_c\right] \mathbf{W}_{c}\right) \\ 
    \alpha_j &= \frac{\exp \left(\mathbf{p}_j \mathbf{W}_{p}\right)}{\sum_{i=1}^{l_e} \exp \left(\mathbf{p}_i \mathbf{W}_{p}\right)} \\
    \mathbf{h}^{g}_{e} &= \sum_{j=1}^{l_{e}} \alpha_{j} \mathbf{H}_{ej}
    \label{eq:attn-e}
\end{align}
where \([\cdot; \cdot]\) denotes the concatenation of two vectors and \(\mathbf{W}_c \in \mathbb{R}^{2d \times d}\) and \(\mathbf{W}_p \in \mathbb{R}^{d \times 1}\) are the trainable parameters. \(l_c\) and \(l_e\) are the length of claim and evidence, respectively. We denote Eq. (\ref{eq:attn-s}) - (\ref{eq:attn-e}) as \(\textbf{ATTN}(\mathbf{H_e}, \mathbf{h}^g_c)\) and the attention modules can be easily extended to multi-head ones by concatenating outputs of each head. It is worth noting that based on the fine-grained representations our graph-based model outputs, the above attention mechanism can be replaced by any interaction method in previous work, which we further discuss in Section \ref{sec:downstream}. 

As \citet{vo2021hierarchical} empirically demonstrate that claim speaker and evidence publisher information is important for verification, we extend claim and evidence representations by concatenating them with corresponding information vectors, i.e., \(\mathbf{h}^{f}_c = [\mathbf{h}^{g}_c; \mathbf{s}]\) and \(\mathbf{h}^{r}_e = [\mathbf{h}^{g}_e; \mathbf{p}]\).

After obtaining the claim and evidence representations, we further employ another attentive network, which is of the same structure as the above, to capture the document-level interaction between a claim and several evidences:

\begin{align}
    \mathbf{H}^{r}_e &= [\mathbf{h}^{r}_{e1}; \mathbf{h}^{r}_{e2}; \ldots; \mathbf{h}^{r}_{en}] \\
    \mathbf{h}^{f}_e &= \textbf{ATTN}(\mathbf{H}^{r}_e, \mathbf{h}^{f}_c)
\end{align}
where \(\mathbf{H}^{r}_e\) denotes the concatenation of embeddings of \(n\) evidences. Eventually, we integrate claim and evidence embeddings into one unified representation via concatenation, followed by a multi-layer perceptron to output the veracity prediction \(\hat{y}\). 
\begin{align}
    \mathbf{h}^f &= [\mathbf{h}^{f}_c; \mathbf{h}^{f}_e] \\
    \hat{y} &= \text{Softmax}(\mathbf{W}_f \mathbf{h}^f + \mathbf{b}_f)
\end{align}

\subsubsection{Training Objective}
As it is fundamentally a classification task, we utilize the standard cross entropy loss as the objective function, which can be written as:
\begin{equation}
    \mathcal{L}_{\Theta}(y, \hat{y})=-(y \log \hat{y}+(1-y) \log (1-\hat{y}))
\end{equation}
where \(y \in \{0, 1\}\) denotes the label of each unverified news.

\section{experiments}
In this section, we conduct comprehensive experiments to answer the following research questions:
\begin{itemize}
    \item RQ1: How does \themodel perform compared to previous fake news detection baselines?
    \item RQ2: How does the redundant information involved in evidences affect the fake news detection?
    \item RQ3: How is the performance of different semantic encoders? 
    \item RQ4: How does \themodel perform with different interaction modules?
    \item RQ5: How does \themodel perform under different hyperparameter settings?
\end{itemize}

\subsection{Experimental Setup}
\subsubsection{Datasets}
We utilize two widely used datasets to verify our proposed model. The detailed statistics is summarized in Table \ref{tab:dataset}.
\begin{itemize}
    \item Snopes \cite{Popat2017WhereTT}. Claims and their corresponding labels (\(true\) or \(false\)) are collected from the fack-checking website\footnote{https://www.snopes.com/}. Taking each claim as a query, the evidences and their publishers are retrieved via the search engine.
    \item PolitiFact \cite{vlachos-riedel-2014-fact}. Claim-label pairs are collected from another fact-checking website\footnote{https://www.politifact.com/} about US politics and evidences are obtained in a similar way to that in Snopes. Aside from publisher information, claim promulgators are added into the dataset. Following previous work \cite{Rashkin2017TruthOV, popat2018declare, vo2021hierarchical}, we merge \(true\), \(mostly\) \(true\), \(half\) \(true\) into the unified class \(true\) and \(false\), \(mostly\) \(false\), \(pants\) \(on\) \(fire\) into \(false\). 
\end{itemize}

\begin{table}[]
  \resizebox{0.48\textwidth}{!}{
    \footnotesize
    \begin{tabular}{cccccc}
        \toprule
        Dataset & \# True & \# False & \# Evi. & \# Spe. & \# Pub. \\
        \midrule
        Snopes & 1164  & 3177  & 29242 & N/A   & 12236 \\
        PolitiFact & 1867  & 1701  & 29556 & 664   & 4542 \\
        \bottomrule
    \end{tabular}
  }
  \caption{The statistics of two datasets. The symbol ``\#'' denotes ``the number of''. ``True'' and ``False'' stand for true claims and false claims, respectively. ``Evi.', `Spe.'', and ``Pub.'' denote evidences, speakers and publishers.}
  \label{tab:dataset}
\end{table}

\begin{table*}[htbp]
    \centering
    \resizebox{\textwidth}{!}{
    \begin{tabular}{ccccccccc|cccccccc}
    \hline
    \multirow{2}{*}{Method} & \multicolumn{8}{c|}{Snopes}                                     & \multicolumn{8}{c}{PolitiFact} \\
    \cline{2-17}
    & F1-Ma & F1-Mi & F1-T    & P-T & R-T & F1-F    & P-F  & R-F & F1-Ma & F1-Mi & F1-T    & P-T & R-T & F1-F    & P-F     & R-F  \\
    \hline
    LSTM    & 0.621    & 0.719      & 0.430     & 0.484    & 0.397   & 0.812   & 0.791   & 0.837  & 0.606   & 0.609    & 0.618      & 0.632   & 0.613      & 0.593   & 0.590   & 0.604 \\
    TextCNN & 0.631    & 0.720      & 0.450     & 0.482    & 0.430   & 0.812   & 0.799   & 0.826 & 0.604   & 0.607    & 0.615      & 0.630   & 0.610      & 0.592   & 0.591   & 0.604 \\
    BERT & 0.621	& 0.716	& 0.431	& 0.477	& 0.407	& 0.810	& 0.793	& 0.830  & 0.597	& 0.598	& 0.608	& 0.619	& 0.599	& 0.586	& 0.577	& 0.597  \\
    \hline
    DeClarE & 0.725      & 0.786      & 0.594    & 0.610    & 0.579    & 0.857   & 0.852  & 0.863 & 0.653   & 0.652     & 0.675    & 0.667   & 0.683      & 0.631   & 0.637   & 0.625 \\
    HAN     & 0.752      & 0.802      & 0.636    & 0.625    & 0.647    & 0.868   & 0.876  & 0.861 & 0.661     & 0.660     & 0.679    & 0.676   & 0.682      & 0.643   & 0.650   & \textbf{0.637} \\
    EHIAN   & 0.784      & 0.828      & 0.684    & 0.617    & 0.768    & 0.885   & 0.882  & 0.890 & 0.676     & 0.679     & 0.689    & 0.686   & 0.693      & 0.655   & 0.675   & 0.636 \\
    MAC     & 0.786      & 0.833      & 0.687    & 0.700    & 0.686    & 0.886   & 0.886  & 0.887 & 0.672     & 0.673     & 0.718    & 0.675   & 0.735      & 0.643   & 0.676   & 0.617 \\
    CICD    & 0.789      & 0.837      & 0.691    & 0.632    & \textbf{0.775}    & 0.893   & 0.890  & 0.895 & 0.682     & 0.685     & 0.702    & \textbf{0.689}   & 0.714      & 0.657   & 0.691   & 0.629 \\
    \hline
    GET     & \(\textbf{0.800}^\ddag\)      & \(\textbf{0.846}^\ddag\)      & \(\textbf{0.705}^\ddag\)    & \(\textbf{0.721}^\ddag\)    & 0.694    & \(\textbf{0.895}^\ddag\)   & \textbf{0.890}  & \(\textbf{0.902}^\ddag\)  & \(\textbf{0.691}^\ddag\)    & \(\textbf{0.694}^\ddag\)    & \(\textbf{0.723}^\ddag\) & 0.687 & \(\textbf{0.764}^\ddag\)      & \(\textbf{0.660}^\ddag\)  & \(\textbf{0.708}^\ddag\)  & 0.629 \\
    \hline
    \end{tabular}
    }
  \caption{The model comparison on two datasets Snopes and PolitiFact. ``F1-Ma'' and ``Fi-Mi'' denote the metrics F1-Macro and F1-Micro, respectively. ``-T'' represents ``True News as Positive'' and ``-F'' denotes ``Fake news as Positive'' in computing the precision and recall values. The best performance is highlighted in boldface. \(\ddag\) indicates that the performance improvement is significant with p-value \(\leq\) 0.05.}
  \label{tab:comp}
\end{table*}%

\subsubsection{Baselines}
To demonstrate the effectiveness of our proposed model \themodel, we compare it with several existing methods, including both pattern- and evidence-based models, the specific description is listed as follows:

\textbf{Pattern-based methods.}
\begin{itemize}
    \item LSTM \cite{lstm}. They utilize LSTM to encode the semantics with the news as input and obtain the final representation of claim via the average pooling. 
    \item TextCNN \cite{textcnn}. They apply a 1D-convolutional network to embed the semantics of claim.
    \item BERT \cite{bert}. They employ BERT to learn the representation of claim. A linear layer is stacked over the special token [CLS] to output the final prediction. 
\end{itemize}

\textbf{Evidence-based methods.}
\begin{itemize}
    \item DeClarE \cite{popat2018declare}. They employ BiLSTMs to embed the semantics of evidences and obtain the claim's representation via average pooling, followed by an attention mechanism performing among claim and each word in evidences to generate the final claim-aware representation.
    \item HAN \cite{ma2019sentence}. They use GRUs to embed semantics and design two modules named topic coherence and semantic entailment to model the claim-evidence interaction, which are based on sentence-level attention mechanism. 
    \item EHIAN \cite{wu2020evidence}. They utilize self-attention mechanism to learn semantics and concentrate on the important part of evidences for interaction.
    \item MAC \cite{vo2021hierarchical}. They introduce a hierarchical attentive framework to model both word- and evidence-level interaction.
    \item CICD \cite{wu2021unified}. They introduce individual and collective cognition view-based interaction to explore both local and global opinions towards a claim.
\end{itemize}

\subsubsection{Implementation Details}
We introduce the specific settings in our experiments including hyperparameters, training settings, and the experimental environment.

Following previous work \cite{popat2018declare, vo2021hierarchical}, we utilize the same data split\footnote{https://github.com/nguyenvo09/EACL2021/tree/main/formatted\_data/declare} to train and test our model. We also report 5-fold cross validation results, where 4 folds are used for training and the rest one fold is for testing. We utilize Adam optimizer with a learning rate \(lr=0.0001\) and weight decay \(decay=0.001\). The model early stops when F1-macro does not increase in 10 epochs and the maximum number of epoch is 100. We set the maximum length of claims and evidences in both datasets as 30 and 100, respectively. The number of evidences \(n = 30\) and the batch size is 32. We set the redundancy discarding rate \(r = 0.4\), i.e., \(k = rl_e\) will be filtered out in a semantic refinement layer, where \(l_e\) is the length of evidence. The number of semantics encoder layer \(T_E = 1\) and evidence semantics miner layer \(T_R = 1\). The number of word-level and document-level attentive readout head as 5 and 2 for Snopes (3 and 1 for PolitiFact), the dimension of publisher and speaker embedding is both 128, following the work \cite{vo2021hierarchical}. We use the Glove pretrained embedding with the dimension \(d=300\) for all baselines for a fair comparison.
We conduct all experiments using PyTorch 1.5.1 on a Linux server equipped with GeForce RTX 3090 GPUs (with 24GB memory each) and AMD EPYC 7742 (256) CPUs.
\subsection{Model Comparison (RQ1)}
We compare our model \themodel with eight baselines\footnote{As some evidence-based methods do not release codes, we reproduce results carefully following settings reported in their original paper.}, including three pattern-based methods and five evidence-based methods. The overall results are shown in Table \ref{tab:comp}, from which we have the following observations:

Firstly, our proposed model \themodel outperforms all existing methods on most of metrics on both two datasets by a significant margin, demonstrating the effectiveness of \themodel. It is worth noting that \themodel stands out from the recent three sequential-based baselines (EHIAN, MAC, and CICD) whose performance is close, indicating the positive impact of introducing graph-based models to evidence-based fake news detection. In detail, compared to the strongest baselines CICD on two datasets, \themodel advances the performance about 1 percent on F1-Macro and F1-Micro, which are the evaluation metrics better reflect the overall detection capability of models. With regard to the more fine-grained evaluation, i.e., `True news as Positive' and `Fake news as Positive', \themodel also achieve the best results on the F1 score on two datasets, where the F1 score is more representative than Precision and Recall since it takes into account both of them synthetically. 

Secondly, compared to the pattern-based methods (i.e., the first three methods in Table \ref{tab:comp}), evidence-based approaches have a substantial performance improvement. This is probably due to the better generalization of evidence-based methods, where the external information is utilized to probe the claim veracity, avoiding the over-reliance on text patterns.
In addition, the performance of BERT is similar to that of other pattern-based approaches. We suspect the reason is probably that claims are short and contain lots of noises (e.g., spelling errors and domain-specific abbreviations), which are rarely appeared in the pretraining corpus, thus it is hard for BERT to transfer the contextual information learned from the pretrained stage. 

Thirdly, among five evidence-based baselines, the performance of DeClarE and HAN is inferior to other three models, which is mainly because they lack exploring the different grain-sized semantics. Specifically, DeClarE only considers word-level semantic interaction and HAN solely relies on document-level representations to model claim-evidence interaction. However, the rest of evidence-based methods all consider multi-level semantics, thus achieving better performance.

\begin{figure}[t]
   \begin{center}
   \includegraphics[width=0.47\textwidth]{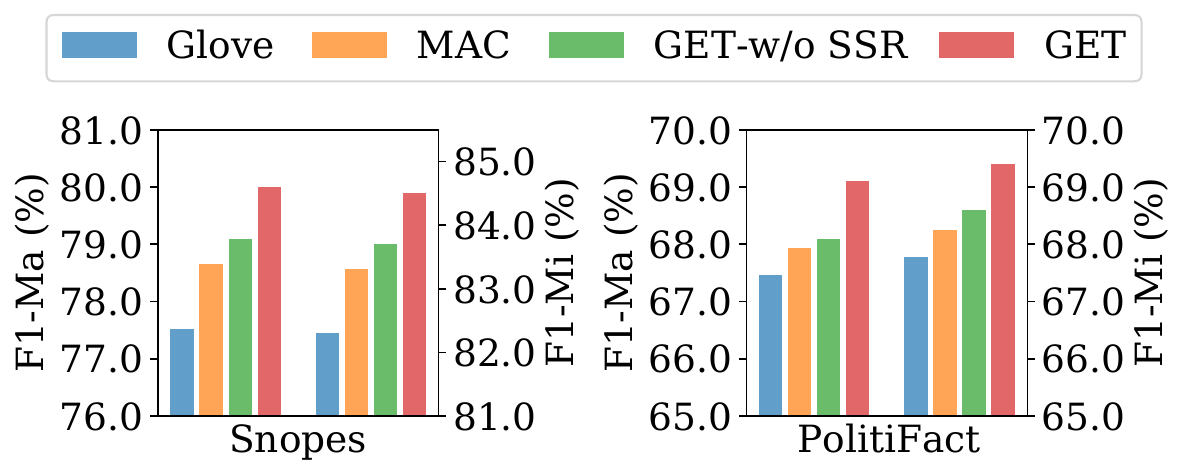}
   \end{center}
   \caption{The performance comparison between \themodel and model variants with different semantic encoders (Glove and MAC) and without structure refinement (\themodel-w/o SSR).}
   \label{fig:ablation study}
\end{figure}

\subsection{Ablation Study (RQ2, RQ3)}

To verify the positive effect of structure refinement for reducing the useless redundancy in evidences, we conduct the ablation study where the structure learning layer is removed and other parts are kept unchanged. We name this model variant as \themodel-w/o SSR. As shown in Figure \ref{fig:ablation study}, we can observe an obvious decline on both datasets regarding the F1-Micro and F1-Macro. This demonstrates the necessity of performing structure refinement on semantic graphs and confirm the effectiveness of our structure learning method. Furthermore, it also indicates that reducing the effect of unimportant information via attention mechanisms will lead to suboptimal results, since they still maintain the noisy semantic structure unchanged \cite{chen2020gsl} (i.e., specifically, all words will participate in the claim-evidence interaction). Therefore, the effect of structure refinement is not overlapped with the attention mechanism, but further goes beyond.  



To demonstrate the superiority of the proposed graph-based semantics encoder, we further conduct experiments on two model variants. One is named Glove, where the pretrained word embeddings are directly fed into the attentive readout layer; the other is named MAC, where the semantics encoder is a BiLSTM the same as the baseline \cite{vo2021hierarchical}. As shown in Figure \ref{fig:ablation study}, Glove has the poorest performance since the contextual information is not captured. Moreover, the performance of \themodel-w/o SSR is superior to that of MAC, indicating that the long-distance structural dependency involved in semantic structure, which is less explored in sequential models, is significant for veracity checking. Note that we choose \themodel-w/o SSR instead of \themodel to be compared with MAC fairly, since the only difference between \themodel-w/o SSR and MAC is the semantics encoder.

\begin{table}[]
  \centering
    \begin{tabular}{ccc >{\columncolor{gray!20}} cc >{\columncolor{gray!20}} c}
    \toprule
    \multicolumn{1}{c}{Dataset} & \multicolumn{1}{c}{Metric} & DeC & \cellcolor{white}\themodel-DeC & EHI & \cellcolor{white}\themodel-EHI \\
    \midrule
    \multirow{4}[1]{*}{Snopes} 
          & F1-Ma & 0.725      & \textbf{0.761} & 0.784      & \textbf{0.795} \\
          & F1-Mi & 0.786      & \textbf{0.813} & 0.828      & \textbf{0.841} \\
          & F1-T  & 0.594      & \textbf{0.649} & 0.684       & \textbf{0.693} \\
          & F1-F  & 0.857      & \textbf{0.873} & 0.885       & \textbf{0.897} \\
    \midrule 
    \multirow{4}[1]{*}{PolitiFact} 
          & F1-Ma & 0.653      & \textbf{0.681} & 0.676      & \textbf{0.688} \\
          & F1-Mi & 0.652      & \textbf{0.685} & 0.679       & \textbf{0.690} \\
          & F1-T  & 0.675      & \textbf{0.714} & 0.689       & \textbf{0.713} \\
          & F1-F  & 0.631      & \textbf{0.647} & 0.655       & \textbf{0.663} \\
    \bottomrule
    \end{tabular}%
  \caption{The performance of \themodel with different claim-evidence interaction modules, compared to their corresponding baselines DeClarE (DeC) and EHIAN (EHI). The superior results are highlighted in boldface.}
  \label{tab:plug-in-play}%
\end{table}%

\subsection{\themodel with Different Claim-evidence Interaction Modules (RQ4)}
\label{sec:downstream}
The \themodel mainly consists of two parts, i.e., graph-based semantic structure mining and attentive graph readout, where the refined semantic structure is obtained in the former stage and the claim-evidence interactions are captured in the latter. As we have mentioned in Section \ref{sec:interaction module}, the semantic structure mining framework can be adaptively connected with any interaction module. Therefore, to further verify the positive impact of graph-based structure mining, we replace the concatenation attention mechanism in our base model with different interaction modules used in existing work. In detail, we choose two modules in representative work: one is the word-level attention mechanism employed in DeClarE \cite{popat2018declare}, the other is the self-attention mechanism utilized to obtain global claim-evidence interactions in EHIAN \cite{wu2020evidence}. We name such two model variants as \themodel-DeC and \themodel-EHI, respectively. Thereafter, we can compare them with DeClarE (DeC) and EHIAN (EHI) to see whether the optimized semantic structure can boost model performance with different downstream interaction modules. 

The experimental results are shown in Table \ref{tab:plug-in-play}. It is obvious that \themodel-DeC and \themodel-EHI both surpass their corresponding competitors, which indicates the effectiveness of our unified graph-based semantic structure mining framework, with being agnostic to the downstream interaction modules. In other words, we can employ such graph-based framework in any evidence-based fake news detection model in a plug-in-play manner, obtaining the fine-grained representations on optimized semantic structures and advancing the model performance.

\subsection{Sensitivity Analysis (RQ5)}
In this section, we conduct experiments to analyse the performance fluctuation of \themodel with respect to different values of key hyperparameters. 

\subsubsection{The number of semantics encoder layer for claims \(T_E\)} 
\begin{figure}[t]
   \begin{center}
   \includegraphics[width=0.47\textwidth]{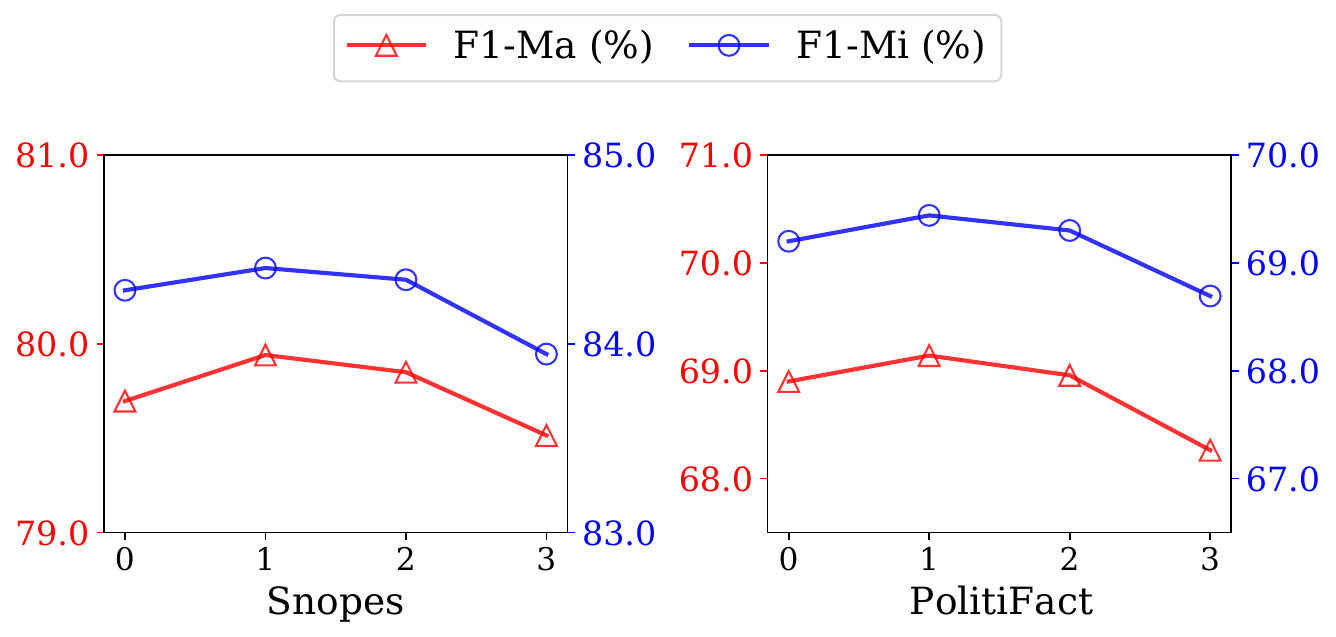}
   \end{center}
   \caption{The influence of different semantics encoder layers \(T_E\) for claims on model performance.}
   \label{fig:ggnn_num}
\end{figure}

This hyperparameter decides propagation field on graphs, since stacking \(T_E\)-layer encoder (GGNN) makes each node aggregate information within \(T_E\)-hop neighborhood. We report the model performance when \(T_E = 0, 1, 2, 3\) (See Figure \ref{fig:ggnn_num}) and summarize the observations as follows:

There is no drastic rise and fall when \(T_E\) is changed from 0 to 3. Specifically, the model with \(T_E = 1\) slightly outperforms its counterparts. We suspect that the close results are due to the short length of claims (the average lengths of claim are about 6 and 8 in Snopes and PolitiFact, respectively), where the semantic structure can be well-explored merely via 1-hop propagation.

Only one obvious decline is observed between \(T_E = 2\) and \(T_E = 3\), which is probably caused by the inappropriate propagation field. When the layer number is 3, each node on graphs aggregate information from 3-hop neighborhood, which may cover all nodes since the claims are short, thus failing to model the local semantic structure and leading to the poor performance.

\begin{figure}[t]
   \begin{center}
   \includegraphics[width=0.47\textwidth]{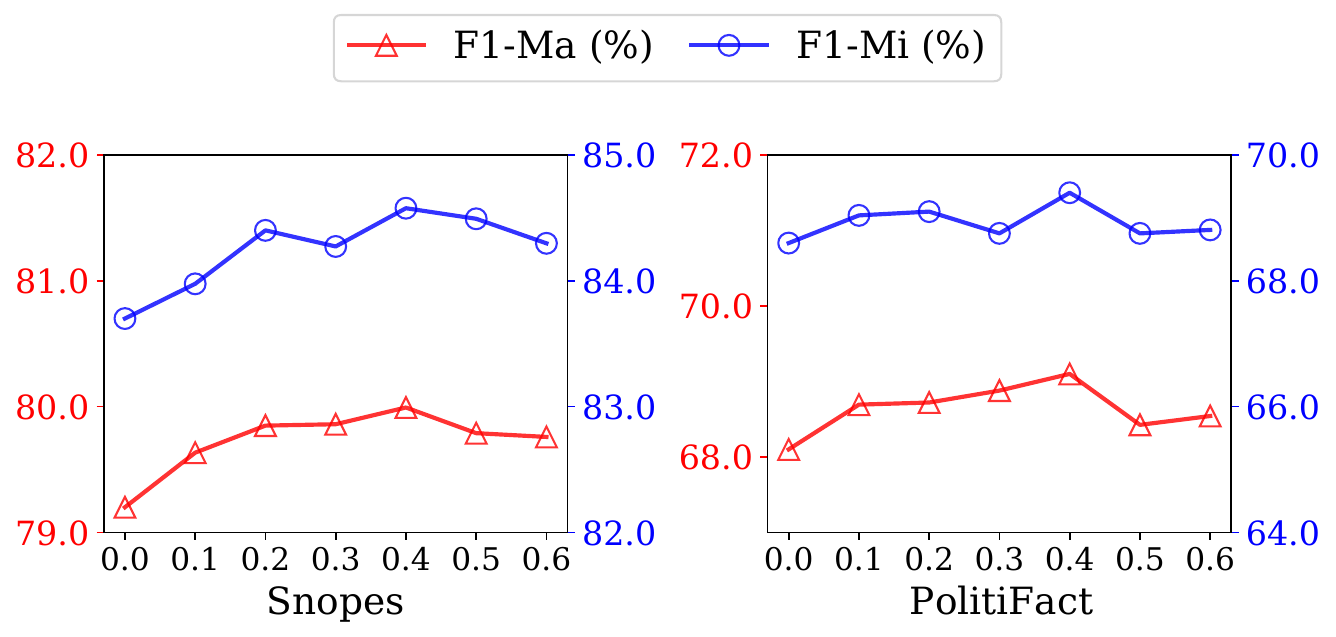}
   \end{center}
   \caption{The influence of different discarding rates \(r\) on model performance.}
   \label{fig:gsl_rate}
\end{figure}

\subsubsection{The discarding rate \(r\)} This rate is also an important hyperparameter in our proposed model \themodel. It decides the proportion of redundant information in evidences we filter out. We test the model with \(r\) ranging from 0 to 0.6 (See Figure \ref{fig:gsl_rate}) and have the following observations:

When \(r = 0\), the model is the same as \themodel-w/o SSR in the ablation study, where structure refinement layer is removed and no words are dropped. We can see that the performance is not satisfactory since redundant information is preserved that may mislead the model.

The performance grows with \(r\) increasing and peaks at the best when \(r = 0.4\), which indicates that reducing redundant information plays a positive role in improving the model performance. When \(r\) is larger than 0.6, a obvious performance decline can be seen. The probable reason is that some useful information for veracity prediction is mistakenly discarded, so that the model fails to capture the rich semantics in evidences, as the \(r\) is too large.


\subsubsection{The number of ESM layer \(T_R\)} 
\begin{figure}[t]
   \begin{center}
   \includegraphics[width=0.47\textwidth]{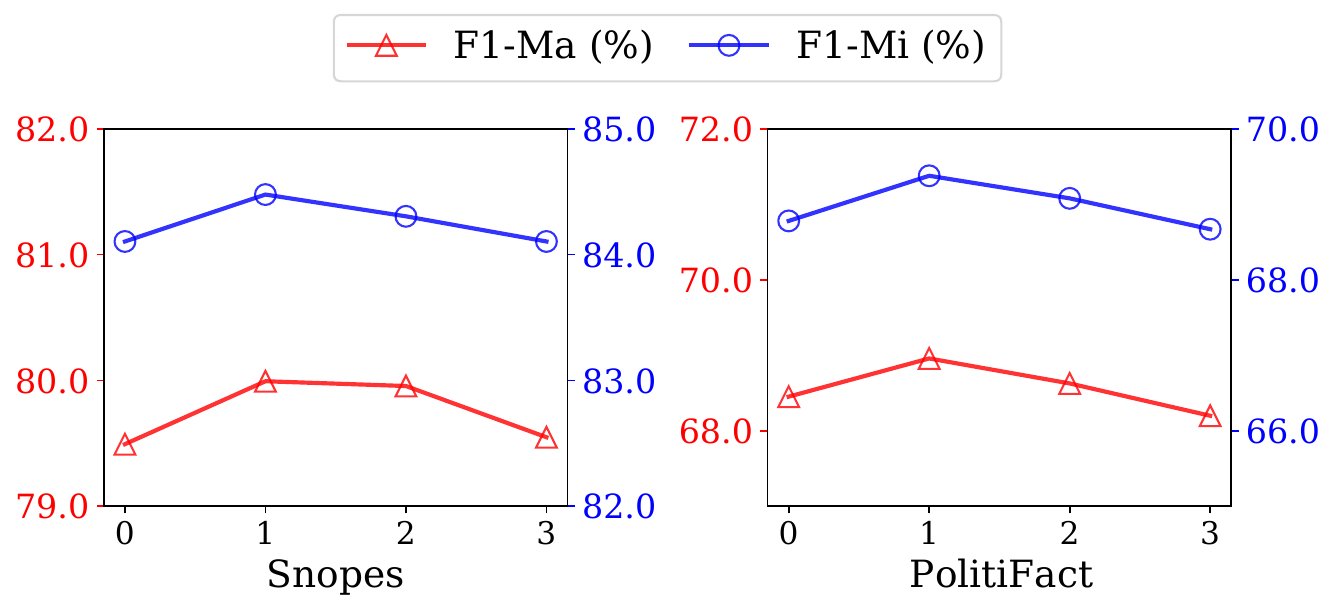}
   \end{center}
   \caption{The influence of different evidence semantics miner layers \(T_R\) on model performance.}
   \label{fig:gsl_num}
\end{figure}

It is a key hyperparameter that controls the information propagation field on graphs and the extent of structure refinement. We observe some phenomena when \(T_R\) increases from 0 to 3 (See Figure \ref{fig:gsl_num}):

The performance is first improved from \(T_R = 0\) to \(T_R = 1\). Note that when \(T_R = 0\), the model downgrades into the one with only a semantics encoder layer. The inferior performance is mainly due to two aspects: 1) It is unable to capture the high-order semantics of long evidences since only features from 1-hop neighborhood are aggregated. 2) Moreover, no redundancy reduction may affect other claim-relevant useful information, since they are fused via neighborhood propagation. Therefore, these drawbacks, in turn, demonstrate the significance of high-order semantics and structure refinement.

A moderate fall of performance can be seen when \(T_R\) ranges from 1 to 3. This is probably because the networks suffer from the over-smoothing problem, which is common in GNNs \cite{Li2018DeeperII}. Besides, the information is overly discarded so that the evidence semantics is not well modeled, which is also a main reason.


\section{conclusion}
In this paper, we have proposed a unified graph-based fake news detection model named \themodel to explore the complex semantic structure. Based on constructed claim and evidence graphs, the long-distance semantic dependency is captured via the information propagation. Moreover, a simple and effective structure learning module is introduced to reduce the redundant information, obtaining fine-grained semantics that are more beneficial for the downstream claim-evidence interaction. We have also validated the performance of \themodel with different interaction methods, where results demonstrate its ability of acting as a plug-in-play module to boost the performance of other fake news detection models.

\begin{acks}
This work is supported by National Natural Science Foundation of China (U19B2038, 61772528, 62141608).
\end{acks}
\bibliographystyle{ACM-Reference-Format}
\bibliography{www22.bib}

\appendix
\section{Visualization of refinement}

\begin{figure}[htbp]
   \begin{center}
   \includegraphics[width=0.47\textwidth]{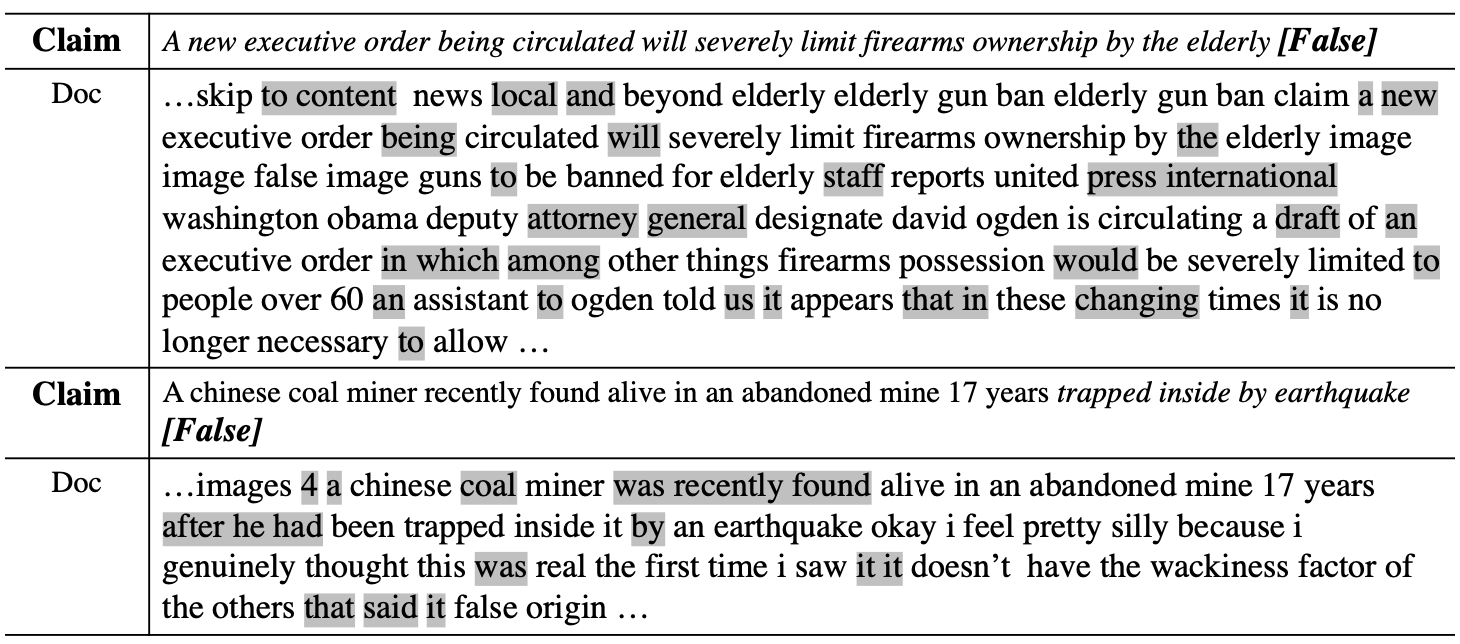}
   \end{center}
   \caption{Visualization of discarded words in the examples in Snopes Dataset. [True/False] indicates veracity of claims.}
   \label{fig:visualization_Snopes}
\end{figure}

\begin{figure}[htbp]
   \begin{center}
   \includegraphics[width=0.47\textwidth]{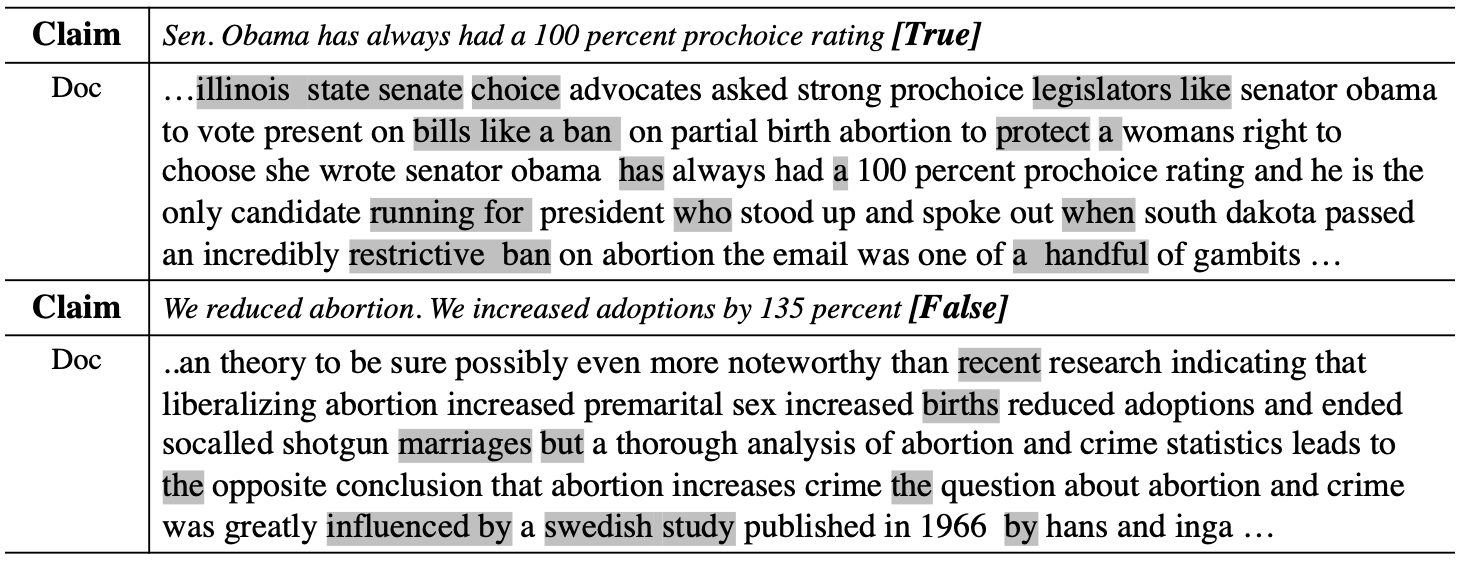}
   \end{center}
   \caption{Visualization of discarded words in the examples in PolitiFact Dataset. [True/False] indicates veracity of claims.}
   \label{fig:visualization_PolitiFact}
\end{figure}

In order to better understand what redundant information is discarded by semantic structure refinement, we visualize examples in both datasets depicted respectively by Figure \ref{fig:visualization_Snopes} and Figure \ref{fig:visualization_PolitiFact}, where the discarded words are highlighted in grey. It is indicated that most dropped words are adverbs, conjunctions, and pronouns which have less valuable information or are relatively unrelated to the news. For instance, words like 'it' and 'by' occur frequently but contribute little to the semantic of text. And in the first example in PoltiFact, several nouns like 'illoinois state senate' have weak connection with its topic and may interfere models' judgement. Therefore, the refinement layer can effectively distill important information and get rid of redundant noises.

\end{document}